\documentclass[conference]{IEEEtran}
\IEEEoverridecommandlockouts
\usepackage{cite}
\usepackage{amsmath,amssymb,amsfonts}
\usepackage{algorithmic}
\usepackage{graphicx}
\usepackage{textcomp}
\usepackage{xcolor}
\usepackage[misc]{ifsym}

\usepackage{bm}
\usepackage[linesnumbered,ruled,vlined]{algorithm2e}
\usepackage{caption}
\usepackage{subcaption}

\def\BibTeX{{\rm B\kern-.05em{\sc i\kern-.025em b}\kern-.08em
    T\kern-.1667em\lower.7ex\hbox{E}\kern-.125emX}}

\usepackage{amsmath,amsfonts,bm}

\def\eqref#1{equation~\ref{#1}}

\def\1{\bm{1}}

\def\vb{{\bm{b}}}

\def\mE{{\bm{E}}}

\def\mG{{\bm{G}}}

\def\mP{{\bm{P}}}

\def\mW{{\bm{W}}}

\DeclareMathAlphabet{\mathsfit}{\encodingdefault}{\sfdefault}{m}{sl}
\SetMathAlphabet{\mathsfit}{bold}{\encodingdefault}{\sfdefault}{bx}{n}

\def\gG{{\mathcal{G}}}

\def\sC{{\mathbb{C}}}

\def\sN{{\mathbb{N}}}

\newcommand{\parents}{Pa} 

\begin{document}

\title{MIPO: Mutual Integration of Patient Journey and Medical Ontology for Healthcare Representation Learning}

\author{
\IEEEauthorblockN{Xueping Peng\IEEEauthorrefmark{1},
        Guodong~Long\IEEEauthorrefmark{1}\textsuperscript{(\Letter)},
        Tao~Shen\IEEEauthorrefmark{1},
        Sen~Wang\IEEEauthorrefmark{2}, Chengqi~Zhang\IEEEauthorrefmark{3}
        Allison~Clarke\IEEEauthorrefmark{4},
        Clement~Schlegel\IEEEauthorrefmark{4}
}
\IEEEauthorblockN{\IEEEauthorrefmark{1}Australian AI Institute,  Faculty of Engineering and IT, University of Technology Sydney, Australia\\
\IEEEauthorrefmark{2}School of Information Technology and Electrical Engineering, The University of Queensland, Australia\\
\IEEEauthorrefmark{3}Department of Data Science and AI, The Hong Kong Polytechnic University, Hong Kong, China\\
\IEEEauthorrefmark{4}Health Economics and Research Division, Department of Health and Aged Care, Australia\\
\textsuperscript{(\Letter)}Corresponding author email: guodong.long@uts.edu.au \\}
}

\maketitle

\begin{abstract}

Representation learning on electronic health records (EHRs) plays a vital role in downstream medical prediction tasks. Although natural language processing techniques, such as recurrent neural networks, and self-attention, have been adapted for learning medical representations from hierarchical, time-stamped EHR data, they often struggle when either general or task-specific data are limited. Recent efforts have attempted to mitigate this challenge by incorporating medical ontologies (i.e., knowledge graphs) into self-supervised tasks like diagnosis prediction. However, two main issues remain: (1) small and uniform ontologies that lack diversity for robust learning, and (2) insufficient attention to the critical contexts or dependencies underlying patient journeys, which could further enhance ontology-based learning. 
To address these gaps, we propose \textbf{MIPO} (Mutual Integration of Patient Journey and Medical Ontology), a robust end-to-end framework that employs a Transformer-based architecture for representation learning. MIPO emphasizes task-specific representation learning through a sequential diagnosis prediction task, while also incorporating an ontology-based disease-typing task. A graph-embedding module is introduced to integrate information from patient visit records, thus alleviating data insufficiency. This setup creates a mutually reinforcing loop, where both patient-journey embedding and ontology embedding benefit from each other. We validate MIPO on two real-world benchmark datasets, showing that it consistently outperforms baseline methods under both sufficient and limited data conditions. Furthermore, the resulting diagnosis embeddings offer improved interpretability, underscoring the promise of MIPO for real-world healthcare applications.
\end{abstract}



\maketitle

\section{Introduction}
Over the past few decades, healthcare information systems have accumulated a considerable amount of electronic health records (EHRs). 
The patient EHRs data typically consists of a sequence of visit records, where each visit includes clinical events such as diagnoses, procedures, medications, and laboratory tests~\cite{Shickel_2018, Song2019-ol}.
Exploiting knowledge from these voluminous EHRs has attracted attention for its potential to benefit patients and caregivers, driving advances in both  academia~\cite{Choi_2016_med2vec,Choi_Bahadori_2017_gram,Ma2018-gu_kame,luo2020hitanet,peng2021sequential,mukherjee2023scope} and industry~\cite{ren2021rapt,Johnson_2016}. 

Recent studies~\cite{Choi_2016_med2vec,zhang2019interpretable,Ma2017-gs_Dipole,Choi_Bahadori_2017_gram,choi2016retain,Ma2018-gu_kame} have explored deep learning techniques to model EHR data for predictive tasks. 
For instance, word embedding techniques, e.g. word2vec \cite{Mikolov_2013_b}, have been adopted in \cite{Choi_2016_med2vec} to learn a vector representation (namely Med2Vec) for each medical concept (e.g. a diagnosis code) from the co-occurrence information without considering the temporal sequential nature of EHR data. 
Furthermore, considering both long-term dependency and sequential information, recurrent neural networks~\cite{Ma2017-gs_Dipole,Choi_Bahadori_2017_gram,choi2016retain,Ma2018-gu_kame}, including LSTM~\cite{hochreiter1997long} and GRU~\cite{cho2014learning}, are used to learn the contextualized representation of EHR data. 
These attempts to learn medical representations are still underperforming for prediction tasks and cannot be practically used for individual patients. 

The reason can be two-fold. The first one is that the squeezing representation capability of EHRs will be bottle-necked for some specific prediction tasks. For example, bringing no more attention to the model when encountering the appearance of terminal disease in a patient's visiting sequences can be detrimental to the performance accuracy in the mortality prediction task. Domain knowledge related to the specific task, with no doubt, can be borrowed to improve performance when modelling. Some NLP learning schemes~\cite{Mikolov_2013,devlin2018bert} have shed light on the extraction of task-wise information for a specific NLP task. Most commonly, they pre-train a general neural module (e.g., word embeddings \cite{Mikolov_2013} and contextual encoder \cite{devlin2018bert}) on a large-scale unlabeled corpus with self-supervised tasks, and then leverage the pre-trained module to initialize the task-specific models for further fine-tuning. The corpus, such as Wikipedia and BookCorpus, consists of words, usually on a scale of billions, so the pre-trained module can produce enough generic representations for efficient fine-tuning convergence and superior performance.  
Unfortunately, compared to the sheer volume of textual data available to NLP tasks, the scale of unlabeled healthcare data is considerably smaller for pre-training to exploit sufficient task-wise medical knowledge.

The second hurdle in the applications of the above models can be blamed on the lack of interpretability, which is considered in priority when making a decision by caregivers.
Graph embedding has been integrated into deep sequence models to improve the performance and provide interpretability. \cite{Choi_Bahadori_2017_gram,Ma2018-gu_kame,wang2024knowledge} train medical code embeddings upon medical ontology by using a graph-based attention mechanism, which delivers a competitive performance with interpretations aligning with medical knowledge.
Note that a strict prerequisite of these works is that each medical code appears as a leaf node in the medical ontology which can be readily satisfied by healthcare data. 
To be clear, medical ontology here refers to a medical knowledge graph, e.g., Clinical Classifications Software (CCS)\footnote{https://www.hcup-us.ahrq.gov/toolssoftware/ccs/ccs.jsp}. 
Despite their success in several healthcare tasks, these methods still labour under two main limitations: 
(1) Unlike factoid knowledge graphs (e.g., Freebase and WikiData), which store hundreds of millions of relational items, the medical ontology contains thousands of diagnosis nodes and merely ``\textit{parent-child}'' hierarchy. Hence, it is insufficient to train expressively powerful code embeddings over the ontology; 
(2) Rich context or dependency information underlying each visit and the patient journey is rarely exploited during medical ontology learning which, however, contains essential information, e.g., complicated diseases. 

To overcome these limitations, we propose a novel and robust healthcare representation learning model, called \textbf{M}utual \textbf{I}ntegration of \textbf{P}atient Journey and Medical \textbf{O}ntology (MIPO). 
It consists of two interactive neural modules: (1) \textit{task-specific representation learning} module and (2) \textit{graph-embedding} module. It aims to infuse medical knowledge into a sequential patient journey by jointly learning the task-specific and the ontology-based objectives.  
To clarify, a \textit{task-specific representation learning} module is composed of two stacked Transformer encoders in a hierarchical scheme. It aims to measure local dependencies among medical codes in each patient visit and further capture long-term dependencies among multiple visits in a patient's journey. 
Concurrently, the \textit{graph-embedding} module learns code embeddings in medical ontology based on both structured knowledges in the graph and contextual information in the patient journey. 
Lastly, we jointly train the model to meet two objectives: one for \textit{task-specific predictive task} based on the representation learning module, and another for \textit{ontology-based disease typing task} based on the graph embedding module. 
Consequently, with such mutual integration and joint learning, MIPO can improve the prediction quality of future diagnoses, guarantee robustness regardless of sufficient or insufficient data, and make the learned patient journeys and diagnoses interpretable. 
Our main contributions are summarized as follows:
\begin{itemize}
	\item We propose MIPO, an end-to-end, novel and robust model to accurately predict patients’ future visit information with mutual integration of patient journey and medical ontology.
	\item We design an \textit{ontology-based disease typing task} in conjunction with the \textit{task-specific predictive task}, to learn effective and robust healthcare representations.
	\item We qualitatively demonstrate the interpretability of the learned representations of medical codes and quantitatively validate the effectiveness of the proposed MIPO.
\end{itemize}


\section{Related Work}\label{sec:rel}
In this section, we review the work related to mining EHRs data with deep learning techniques, especially for diagnosis prediction. We then introduce some of the latest work on the transformer-based model with knowledge graph.

\subsection{Deep Learning for EHRs Data}
In recent years, researchers have proposed various deep learning models to garner knowledge from massive EHRs and shown their superior ability in medical event predictions~\cite{Ma2017-gs_Dipole,li2020behrt,li2019fine,gao2020stagenet,bai2019improving,song2018attend,zeng2020multilevel,peng2021sequential,ren2021rapt,luo2020hitanet}. 
Previous studies recommend using recurrent neural networks (RNNs) for patient subtyping~\cite{baytas2017patient,gao2020stagenet}, modelling  disease progression~\cite{pham2016deepcare}, and time-series healthcare-data analysis~\cite{ruan2019representation}. Convolutional neural networks (CNNs) are exploited to predict unplanned readmission~\cite{nguyen2016deepr} and risk\cite{Ma2018-ao} with EHRs. Stacked autoencoders are employed to generate sequential EHRs data~\cite{lee2020generating}. The emerging transformer-based BERT model is used for acquiring knowledge from clinical notes~\cite{li2019fine} and future visit prediction~\cite{li2020behrt}.

Diagnosis prediction is an important application in healthcare analytics~\cite{Choi_Bahadori_2017_gram,Ma2018-gu_kame,niu2021label,zang2021scehr,peng2019attentive}, 
which leverages a patient's sequential visit records to predict future visit information. RETAIN~\cite{choi2016retain} and Dipole~\cite{Ma2017-gs_Dipole} are two representative RNN-based diagnoses predictive models. RETAIN employs RNNs to model reverse, time-ordered-EHRs sequential visits with an attention mechanism for the binary prediction tasks.
Dipole applies Bi-LSTM and attention mechanisms to predict patient visit information, which enhances the temporal data modelling  ability of predictive models. 
However, those approaches could suffer from a data insufficiency~\cite{Choi_Bahadori_2017_gram}. To
alleviate this problem; GRAM~\cite{Choi_Bahadori_2017_gram} and KAME~\cite{Ma2018-gu_kame} exploit the information from external medical knowledge graph to learn robust representations and an RNN to model patient visits. Although this achieves state-of-the-art performance, both models lack effective aggregation of multiple medical codes in a visit and, heterogeneous information integration of patient journey and knowledge graph, which should be taken as an advantage for improving performance.

\subsection{Transformer-based Model with Knowledge Graph}
Devlin et al.~\cite{devlin2018bert} propose a deep bi-directional model with multiple-layer Transformers (BERT), which achieves the state-of-the-art results for various NLP tasks  (eg, question answering, named entity recognition, and relation extraction). 
The latest, transformer-based BEHRT~\cite{li2020behrt} makes direct use of the original BERT~\cite{devlin2018bert} to model the patient's sequential EHRs data by taking each visit as a sentence and each medical concept as a word to predict future visit information. 
BioBERT~\cite{lee2020biobert}, ClinicalBERT~\cite{huang2019clinicalbert} and Med-BERT~\cite{rasmy2021med} achieve new state-of-the-art results on various biomedical NLP tasks through simple fine-tuning techniques with medical corpus. Furthermore, ERNIE~\cite{zhang2019ernie}, K-BERT~\cite{liu2019kbert} and KGET~\cite{wang2024knowledge} infuse a knowledge graph into pre-trained BERT to further enhance language representation. As EHRs data have different characteristics to natural language, (e.g., medical code can provide a one-to-one map node of knowledge graph), medical codes are time-ordered in a visit. Thus, enhanced language models, such as ERNIE and K-BERT, cannot apply directly to tackling healthcare problems. However, the ideas from pairing the NLP language models with knowledge graphs motivate us to propose MIPO to mutually integrate patient journey and medical knowledge for healthcare representation learning. 

\begin{figure}[th]
  \includegraphics[width=0.47\textwidth]{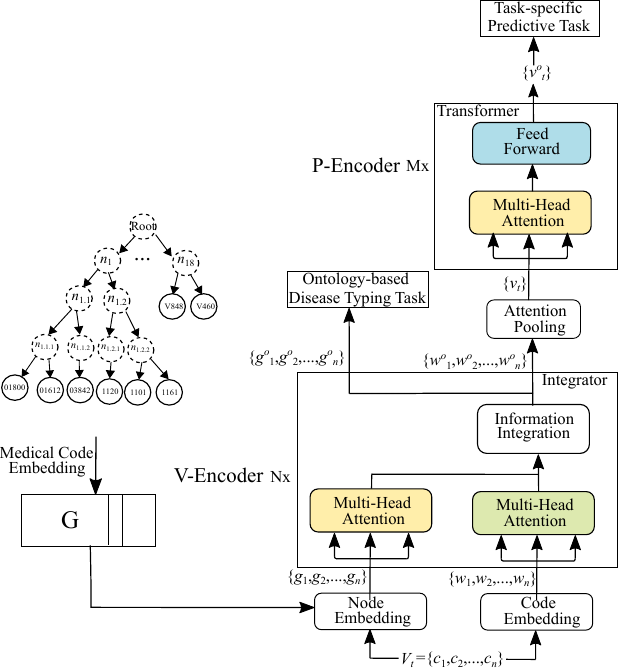}
  \caption{The model architecture of MIPO. The graph is formatted as a hierarchical tree, in which, the root node is virtual. To construct the tree, the leaf nodes (solid circles) denote fine-grained diagnoses, and the non-leaf nodes (dotted circles) denote coarse-grained disease concepts.}
  \label{fig:model1}
\end{figure}


\section{Methodology}\label{sec:Method}

\subsection{Notations}\label{sec:nots}

We denote the set of medical codes from the EHRs data as $c_1, c_2, \dots, c_{|\mathbb{C}|} \in \displaystyle \sC$ and $|\displaystyle \sC|$ is the number of unique medical codes. Patients' clinical records can be represented by a sequence of visits $\displaystyle \mP=\langle V_1, \dots, V_t, \dots, V_T\rangle$,  which is referred to as the patient journey in the paper, where $T$ is the visit number in the patient journey. And each visit $V_t$ consists of a subset of medical codes ($V_t \subseteq \displaystyle \sC$). For clear demonstration, all algorithms will be presented with a single patient's journey.
On the other hand, a medical ontology $\displaystyle \gG$ contains the hierarchy of various medical concepts with the \textit{parent-child} semantic relationship, which is a well-organized ontology in healthcare. In particular, the medical ontology $\displaystyle \gG$ is a directed acyclic graph (DAG) and the nodes of $\displaystyle \gG$ consist of leaves and their ancestors. 
Each leaf node refers to a medical code in $\displaystyle \sC$, which is associated with a sequence of ancestors from the leaf to the root of $\displaystyle \gG$. And
each ancestor node belongs to the set $\displaystyle \sN = {n_{|\mathbb{C}|+1}, n_{|\mathbb{C}|+2}, \dots ,n_{|\mathbb{C}|+|\mathbb{N}|}}$, where $|\displaystyle \sN|$ is the number of ancestor codes in $\displaystyle \gG$. A parent in the knowledge graph $\displaystyle \gG$ represents a related but more general concept over its children. 

\subsection{Model Architecture}

As shown in Figure~\ref{fig:model1}, the whole model architecture of MIPO consists of an embedded knowledge graph and two stacked modules and an attention pooling layer. Using given knowledge graph $\displaystyle \gG$, we can obtain the embedding matrix $\displaystyle \mG$ of medical codes with graph-based attention mechanism~\cite{Choi_Bahadori_2017_gram}. Given the \textit{t}-th visit information of a patient $V_t$, each medical code corresponding to a leaf node in $\displaystyle \mG$ in $V_t$ is embedded into a vector representation with the learned $\displaystyle \mG$, which imposes medical knowledge on the network architecture. The two stacked modules are: (1) the underlying knowledgeable encoder ($\texttt{V-Encoder}$) responsible for integrating extra medical knowledge information into visit information from basic embedding, so that we can represent the heterogeneous information of medical codes and graph nodes into a united feature space, and (2) the upper patient encoder ($\texttt{P-Encoder}$) is responsible of capturing contextual and sequential information from the underlying layer. Attention Pooling~\cite{lin2017structured,liu2016learning} explores the importance of each code within an entire visit. 
It works by compressing a set of medical code embeddings from a visit into a single context-aware vector representation for the upper $\texttt{P-Encoder}$. 
We also denote the number of $\texttt{V-Encoder}$ layers as \textit{N}, and the number $\texttt{P-Encoder}$ layers as \textit{M}. The output of $\texttt{P-Encoder}$ is used to predict the information of next visit.

To be specific, given a patient's visit $V_t=\{c_1, c_2, \dots, c_n\}$, where \textit{n} is the number of medical codes in the visit, we can first obtain its corresponding basic code embedding $\{\bm{w}_1, \bm{w}_2, \dots, \bm{w}_n\}$ via code embedding layer and its node embedding $\{\bm{g}_1, \bm{g}_2, \dots, \bm{g}_n\}$ via learned knowledge graph embedding matrix \textbf{G}. More details of the knowledge graph embedding are introduced in Section~\ref{sec:kg}.  Then, MIPO adopts a knowledgeable encoder $\texttt{V-Encoder}$ to inject the medical knowledge information into healthcare representation, where both $\{\bm{w}_1, \bm{w}_2, \dots, \bm{w}_n\}$ and $\{\bm{g}_1, \bm{g}_2, \dots, \bm{g}_n\}$ are fed into $\texttt{V-Encoder}$ for fusing heterogeneous information and computing final output embeddings, 
\begin{equation}\label{eq:V-Transformer}
\begin{aligned}
\{\bm{w}_1^{o}, &  \bm{w}^o_2, \ldots, \bm{w}_n^{o} \},   \{\bm{g}^o_1, \bm{g}^o_2, \dots, \bm{g}^o_n\} = \texttt{V-Encoder}( \\ 
& \{\bm{w}_1, \bm{w}_2, \ldots, \bm{w}_n\},  \{\bm{g}_1, \bm{g}_2, \dots, \bm{g}_n\}),
\end{aligned}
\end{equation}
where $\{\bm{g}_1^{o}, \ldots, \bm{g}_n^{o}\}$ will be used as features for ontology-based disease typing task and $\{\bm{w}_1^{o}, \bm{w}_2, \ldots, \bm{w}_n^{o} \}$ will fed into upper attention pooling layer (detailed in Section~\ref{sec:attn_pool}) to compress then to a vector $\bm{v}_t$ representing the visit. More details of the knowledgeable encoder $\texttt{V-Encoder}$ will be introduced in Section~\ref{sec:k-encoder}. 

After computing $\bm{v}_t$, MIPO employs a $\texttt{P-Encoder}$ to capture the contextual and sequential information, $\bm{v}^o_t = \texttt{P-Encoder}(\bm{v}_t)$, where $\texttt{P-Encoder}(\cdot)$ is a multi-layer bidirectional Transformer encoder. Since $\texttt{P-Encoder}$ is identical to its implementation in BERT, we exclude a comprehensive description of this module and refer readers to~\cite{devlin2018bert} and~\cite{vaswani2017attention}. 

For simplicity, we take only one patient's visit $V_t$ as an example. However, since most patients have multiple visits to hospital, inputs and outputs of $\texttt{V-Encoder}$ and $\texttt{P-Encoder}$ are multiple visits, e.g., 
\begin{equation}\label{eq:p-encoder}
\{\bm{v}^o_1, \bm{v}^o_2, \ldots, \bm{v}^o_{T-1}\} = \texttt{P-Encoder}(\{\bm{v}_1, \bm{v}_2, \ldots, \bm{v}_{T-1}\}),
\end{equation}
where $\{\bm{v}^o_1, \bm{v}^o_2, \ldots, \bm{v}^o_{T-1}\}$ will be used as features for the task of sequential diseases prediction.

\subsection{Knowledge Graph Embedding}\label{sec:kg}

To mitigate the problem of data insufficiency in healthcare and to learn knowledgeable and generalized representations of medical codes, we employ the attention-based graph embedding approach GRAM~\cite{Choi_Bahadori_2017_gram}. In the medical ontology $\displaystyle \gG$, each leaf node $c_i$ has a basic learnable embedding vector $\displaystyle \mE_{i,:} \in \mathbb{R}^d$ , where $1 \leq i \leq |\displaystyle \sC|$, and \textit{d} represent the dimensionality. And each ancestor code $n_i$ also has an embedding vector 
$\displaystyle \mE_{i,:} \in \mathbb{R}^d$, 
where $|\displaystyle \sC|+1 \leq i \leq |\displaystyle \sC|+|\displaystyle \sN|$. The attention-based graph embedding uses an attention mechanism to learn the \textit{d}-dimensional final embedding $\displaystyle \mG$ of each leaf node \textit{i} (medical code) via:
\begin{equation}\label{eq:g-embedding}
  \displaystyle \mG_{i,:}=\sum_{j\in \displaystyle \parents_\gG(i)}\alpha_{ij}\displaystyle \mE_{j,:}
\end{equation}
where $\displaystyle \parents_\gG(i)$ denotes the set comprised of leaf node \textit{i} and all its ancestors, $\displaystyle \mE_{j,:}$ is the \textit{d}-dimensional basic embedding of the node \textit{j} and $\alpha_{ij}$ is the attention weight on the embedding $\displaystyle \mE_{j,:}$ when calculating $\displaystyle \mG_{i,:}$, which is formulated by following the Softmax function,
\begin{equation}
\alpha_{ij} = \frac{\exp(f(\displaystyle \mE_{i,:}, \displaystyle \mE_{j,:}))}{\sum_{k\in \displaystyle \parents_\gG(i)}\exp(f(\displaystyle \mE_{i,:}, \displaystyle \mE_{k,:}))}.
\end{equation}
\begin{equation}
f(\displaystyle \mE_{i,:}, \displaystyle \mE_{j,:}) = \bm{w}_{\alpha}^T \texttt{tanh} \left(\displaystyle \mW_{\alpha}[\displaystyle \mE_{i,:}; \displaystyle \mE_{j,:}] + \displaystyle \vb_{\alpha}\right),
\end{equation}
where $[\displaystyle \mE_{i,:}; \displaystyle \mE_{j,:}]$ is to concatenate $\displaystyle \mE_{i,:}$ and $\displaystyle \mE_{j,:}$ in the child-ancestor order, $\bm{w}_{\alpha}$, $\displaystyle \mW_{\alpha}$ and $\displaystyle \vb_{\alpha}$ are learnable parameters.

\subsection{Knowledgeable Encoder}\label{sec:k-encoder}
The design of the integrator is inspired by NLP language modelling ERNIE~\cite{zhang2019ernie}. 
However, our proposed model is distinct from ERNIE in three aspects: 1) node embeddings of a knowledge graph is part of our end-to-end MIPO model, while ERNIE uses pre-trained entity embedding from a knowledge graph by TransE~\cite{bordes2013translating}; 2) MIPO has a hierarchical structure, where the underlying knowledgeable encoder is for medical code level and the upper patient encoder is for visit level, while ERNIE is a derivative of BERT, which has not such structure; 3) MIPO aims to improve the predictive  performance with the given knowledge graph as supplementary information.

In the \textit{i}-th integrator, the input code embeddings $\{\bm{w}_1, \bm{w}_2, \ldots, \bm{w}_n\}$ and node embedding $\{\bm{g}_1, \bm{g}_2, \dots, \bm{g}_n\}$ are fed into two different multi-head self-attentions (MSA)~\cite{vaswani2017attention}.
\begin{equation}
\label{e_code2visit}
\small
\begin{aligned}
\{\tilde{\bm{w}}^{(i)}_1, \tilde{\bm{w}}^{(i)}_2, \ldots, \tilde{\bm{w}}^{(i)}_n\} = \texttt{MSA}(\{\bm{w}^{(i-1)}_1, \bm{w}^{(i-1)}_2, \ldots, \bm{w}^{(i-1)}_n\}),\\
\{\tilde{\bm{g}}^{(i)}_1, \tilde{\bm{g}}^{(i)}_2, \ldots, \tilde{\bm{g}}^{(i)}_n\} = \texttt{MSA}(\{\bm{g}^{(i-1)}_1, \bm{g}^{(i-1)}_2, \ldots, \bm{g}^{(i-1)}_n\}).
\end{aligned}
\end{equation}

Then, the \textit{i}-th integrator adopts an information integration layer for the mutual integration of the code and node embedding in a visit, and computes the output embedding for each code and node. For a code $\bm{w}_j$ and its corresponding node $\bm{g}_j$, the information integration process is as follows,
\begin{equation}
\begin{aligned}
	\bm{h}_j &= \sigma (\bm{\tilde{W}}_{c}^{(i)} \bm{\tilde{w}}^{(i)}_j + \bm{\tilde{W}}_{g}^{(i)} \bm{\tilde{g}}^{(i)}_j + \bm{\tilde{b}}^{(i)}),\\
	\bm{w}^{(i)}_j &= \sigma (\bm{W}_{c}^{(i)} \bm{h}_j + \bm{b}^{(i)}_{t}), \\
	\bm{g}^{(i)}_j &= \sigma (\bm{W}_{g}^{(i)} \bm{h}_j + \bm{b}^{(i)}_{e}). \\
\end{aligned}
\end{equation}
where $\bm{h}_j$ is the inner hidden state integrating the information of both the code and the node. $\sigma(\cdot)$ is the non-linear activation function, which is usually the ReLU function.

For simplicity, the $i$-th integrator operation is denoted as follows,
\begin{equation}
\begin{aligned}
\{\bm{w}^{(i)}_1, & \ldots, \bm{w}^{(i)}_n\},\{\bm{g}^{(i)}_1, \ldots, \bm{g}^{(i)}_n\} = \texttt{Integrator} ( \\
&\{\bm{w}^{(i-1)}_1, \ldots, \bm{w}^{(i-1)}_n\},\{\bm{g}^{(i-1)}_1, \ldots, \bm{g}^{(i-1)}_n\}). 
\end{aligned}
\end{equation}

The output embeddings of codes will be used by following attention pooling to compress a set of codes in a visit to a vector, with the output embeddings of nodes used to guarantee the proposed model can learn the reasonable knowledge from given medical ontology.

Note that we exclude position embedding in $\texttt{V-Encoder}$, as medical codes in a visit are not time-ordered.

\subsection{Attention Pooling}\label{sec:attn_pool}
Attention Pooling~\cite{lin2017structured,liu2016learning} explores the importance of each individual code within a patient visit. It works by compressing a set of medical code embeddings from a patient visit into a single context-aware vector representation. Formally, it is written as, 
\begin{equation}
f(\bm{w}^o_i) = w^T \sigma (W^{(1)}\bm{w}^o_i + b^{(1)})+b,
\end{equation}
where $\bm{w}^o_i$ ($1 \leq i \leq n$ ) is one output of $\texttt{V-Encoder}$. The probability distribution is formalized as
\begin{equation}
\label{alpa-add}
\bm{\alpha} = \texttt{softmax}([f(\bm{w^o}_i)]_{i=1}^n).
\end{equation}

The final output $\bm{v}$ of the attention pooling is the weighted average of sampling a code according to its importance, i.e., 
\begin{equation}
\label{eq:v-embedding}
\bm{v} = \sum_{i=1}^n \bm{\alpha} \odot [\bm{w^o}_i]_{i=1}^n.
\end{equation}

\subsection{Learning Healthcare Representation with Predictive Tasks}

We jointly train the MIPO model with a \textit{task-specific predictive task} and an  \textit{ontology-based disease typing task}, such that the mutual integration of knowledge graph and patient journey improves the performance of the healthcare representation learning. 

\subsubsection{Task-specific Predictive Task}
Given a patient's visit records $\bm{P} = \{V_1, V_2, \ldots, V_{T-1}\}$, to capture the EHRs sequential visit behaviour information, we perform the sequential diagnoses predictive task with the objective of predicting the disease codes of the next visit $V_{t}$,  which can be expressed as follows,
\begin{align} 
\label{eq:seq_predict}
 \hat{\bm{y}}^P_{t-1} = \hat{\bm{v}}_{t} = \texttt{Softmax}(\bm{W}_P\bm{v}^o_{t-1} + \bm{b}_P),
\end{align}
\begin{equation}
\label{eq:seq_loss}
\begin{aligned}
& \mathcal{L}_P(V_1, \ldots, V_{T}) = 
\\ & \frac{1}{T-1}\sum_{t=1}^{T-1}\left({\bm{y}^P_t}^{\texttt{T}}\log{\hat{\bm{y}}}^P_t+(1-{\bm{y}^P_t})^{\texttt{T}}\log{(1-\hat{\bm{y}}^P_t)}\right). 
\end{aligned}
\end{equation}
where $\bm{v}^o_{t-1}\in \mathbb{R}^d$ is the output of $\texttt{P-Encoder}$ to denote the representation of the ($t-1$)-th visit, $\bm{W}_P \in \mathbb{R}^{|\mathcal{C}|\times d}$ and $\bm{b}_P \in \mathbb{R}^{|\mathcal{C}|}$ are the learnable parameters. 

\subsubsection{Ontology-based Disease Typing Task}
To Enable MIPO to inject knowledge into healthcare representation by informative graph, we design the task using the output node embeddings of the knowledgeable encoder $\texttt{V-Encoder}$. This task is a multi-label prediction task.
In particular, the non-leaf nodes located from the second layer in medical ontology $\mathcal{G}$ are also known as the disease categories (or types), and each fine-grained diagnosis corresponds to the only disease category by finding its ancestor in the second layer.

As mentioned in Section~\ref{sec:nots}, knowledge graph $\mathcal{G}$ contains the hierarchy of various medical concepts with the \textit{parent-child} semantic relationship, and the medical codes $\displaystyle \sC$ come form its leaf nodes. Ideally, the disease categories in $\mathcal{G}$ will acquire knowledge from the leaf nodes and represent more general medical concepts. Thus, we use the disease categories as targets of the task and the output embeddings of nodes of $\texttt{V-Encoder}$ as input. To be specific, given the codes $V_t = \{c_1, c_2, \dots, c_n\}$ in a visit $V_t$, and its corresponding disease categories $\{n_1, n_2, \ldots ,n_m\}$ (shown in Figure~\ref{fig:model1}) in multi-level hierarchy $\mathcal{G}$, where $m=18$ for CCS Multi-level ontology, we define the disease categories distribution for the medical code $c_i$ in $V_t$ as follows,
\begin{align} 
\label{eq:node_prediction}
\hat{\bm{y}}^V_{t,i} = \texttt{Softmax}(\bm{W}_V\bm{g}^o_{t,i} + \bm{b}_V),
\end{align}
where $\bm{g}^o_{t,i}\in \mathbb{R}^d$ is the output of $\texttt{V-Encoder}$, and $t$ the $t$-th visit, $i$ is the $i$-th code in $t$-th visit, $\bm{W}_V \in \mathbb{R}^{m \times d}$ and $\bm{b}_V \in \mathbb{R}^{m}$ are the learnable parameters. 

Based on Equation~\ref{eq:node_prediction}, we use the cross-entropy between the ground truth visit $\bm{y}^V_{t,i}$ and the predicted visit $\hat{\bm{y}}^V_{t,i}$ to calculate the loss for each
medical code from all the timestamps as follows:
\begin{equation}
\label{eq:node_loss}
\begin{aligned}
& \mathcal{L}_V(V_1, \ldots, V_{T-1}) = 
\\ & \frac{1}{n(T-1)}\sum_{t=1}^{T-1}\sum_{i=1}^{n}\left({\bm{y}^V_{t,i}}^{\texttt{T}}\log{\hat{\bm{y}}}^V_{t,i}+(1-{\bm{y}^V_{t,i}})^{\texttt{T}}\log{(1-\hat{\bm{y}}^V_{t,i})}\right). 
\end{aligned}
\end{equation}
where $T-1$ is the number of the patient's visits, and $n$ is the number of medical codes in a visit.

\subsubsection{Objective Function}
In order to take advantage of the mutual integration of informative knowledge graph and sequential patient journey, we train the two tasks together to improve the performance of the healthcare representation learning, which can be formulated as follows,
\begin{equation}
\label{eq:total_loss}
\mathcal{L}(V_1, \ldots, V_{T}) = \mathcal{L}_P(V_1, \ldots, V_{T}) + \mathcal{L}_V(V_1, \ldots, V_{T-1}). 
\end{equation}

Note that in our implementation, we take the average of the individual cross entropy error for multiple patients. 
Algorithm~\ref{alg} describes the overall training procedure of the proposed MIPO with one individual patient journey.

\begin{algorithm}[ht]
\KwIn{ Medical knowledge graph $\mathcal{G}$, the set of medical codes $\mathcal{C}$ and Patient records $\bm{P} = \{V_1, V_2, \dots, V_{T-1}\}$}

Initialize medical code embedding matrix $\bm{W}$\;
Knowledge graph embedding matrix $\bm{G}$ via Eq.~\ref{eq:g-embedding}\;
Initialize $\texttt{v-list}$ $\bm{to}$ None and $\hat{\bm{y}}^K\texttt{-list}$ $\bm{to}$ None \;
\For{$t \gets 1$ \textbf{to} $(T-1)$}{
  $\bm{W}_t = \{\bm{w}_1, \bm{w}_2, \ldots, \bm{w}_n\}$ \# medical code embedding\;
  $\bm{G}_t = \{\bm{g}_1, \bm{g}_2, \ldots, \bm{g}_n\}$ \# graph node embedding\;
  $ \{\bm{w}_1^o, \bm{w}_2^o, \ldots, \bm{w}_n^o\}, \{\bm{g}^o_1, \bm{g}^o_2, \ldots, \bm{g}^o_n\} = \texttt{V-Encoder}(\bm{W}_t, \bm{G}_t)$ via Eq.~\ref{eq:V-Transformer}\;
  $\bm{v}_t = \texttt{Att-Pool}(\{\bm{w}_1^o, \bm{w}_2^o, \ldots, \bm{w}_n^o\})$ via Eq.~\ref{eq:v-embedding} \# $t$-th visit representation\;
  Add $\bm{v}_t$ $\bm{to}$ $\texttt{v-list}$\;
  Compute predicted first-level category $\hat{\bm{y}}^K_{t,\cdot}$ via Eq.~\ref{eq:node_prediction};
  Add $\hat{\bm{y}}^K_{t,\cdot}$ $\bm{to}$ $\hat{\bm{y}}^K\texttt{-list}$\;
}
$\{\bm{v}^o_1, \bm{v}^o_2, \ldots, \bm{v}^o_{T-1}\} =  \texttt{P-Encoder}(\texttt{v-list})$ via Eq.~\ref{eq:p-encoder}\;
 Compute predicted sequential diagnoses $\hat{\bm{y}}^P$ via Eq.~\ref{eq:seq_predict}\;
 Update the model’s parameters by optimizing the loss via Eq.~\ref{eq:total_loss} using $\hat{\bm{y}}^P$ and $\hat{\bm{y}}^K\texttt{-list}$.
 \caption{The MIPO model}\label{alg}
\end{algorithm}

\section{Experiments}\label{sec:Experim}


\subsection{ Data Description}


\subsubsection{MIMIC-III Dataset}
The MIMIC-III dataset~\cite{Johnson_2016} is an open-source, large-scale, de-identified dataset of ICU patients and their EHRs. The diagnosis codes in the dataset follow the ICD9 standard. The dataset consists of medical records of 7,499 intensive care unit (ICU) patients over 11 years, where we chose patients who had made at least two visits. We use MIMIC to represent MIMIC-III in the experiment.

\subsubsection{eICU Dataset}

The eICU dataset~\cite{pollard2018eicu} is another publicly available EHRs dataset, which is a multi-center database comprising de-identified health data associated with over 200,000 admissions to ICUs across the United States between 2014-2015. The dataset consists of medical records of 16,180 ICU patients, where we follow MIMIC to choose patients who had made at least two visits.

\begin{table}
\centering
  \caption{Statistics of the datasets.}
  \label{tab:stats}
  \begin{tabular}{lcc}
    \hline
    Dataset&MIMIC&eICU\\
    \hline
    \# of patients & 7,499&16,180 \\
    \# of visits &  19,911&39,912 \\
    Avg. \# of visits per patient & 2.66&2.47 \\
   \hline
    \# of unique ICD9 codes &  4,880&758 \\
    Avg. \# of ICD9 codes per visit & 13.06&5.21\\
    Max \# of ICD9 codes per visit& 39&57\\
    \hline
    \# of category codes&272 &167\\
    Avg. \# of category codes per visit&11.23&4.72\\
    Max \# of category codes per visit&34&33\\
    \hline
    \# of disease typing code& 18 &18\\
    Avg. \# of disease typing codes per visit&6.57&3.42\\
    Max \# of disease typing codes per visit&15&14\\
  \hline
\end{tabular}
\end{table}

Table~\ref{tab:stats} shows the statistical details about the two datasets. As the table shown, those two representative datasets can be used to extensively evaluate different aspects of the models. The number of patients and visits in the eICU dataset is big enough to validate the performance of the proposed MIPO with long visit records. The MIMIC dataset consists of very short visits, and the number of patients is smaller. 
With these two different types of datasets, we can fully and correctly validate the performance of all the diagnosis prediction approaches.

\begin{table*}[t]
	\caption{ Performance comparison of sequential diagnoses prediction.}\label{tab:Accuracy}
	\centering
	\scalebox{0.96}{%
		\begin{tabular}{llccccccccccccc}
			\hline
			Dataset & Model&\multicolumn{6}{c}{Prec@k}&\multicolumn{6}{c}{Acc@k} \\
			\cline{3-8} \cline{10-15} & &5 & 10 &15& 20 &25& 30&  &5 & 10 &15& 20 &25& 30 \\ \hline
			&BRNN     & 
			0.5707&0.5112&0.5270&0.5718&0.6234&0.6690& & 
			0.2692&0.4028&0.4933&0.5636&0.6220&0.6690 \\ 
			&RETAIN    & 
			0.5769&0.5071&0.5280&0.5700&0.6214&0.6721& & 
			0.2721&0.3976&0.4936&0.5617&0.6201&0.6721\\ 
			&Dipole  & 
			0.5750&0.5104&0.5334&0.5813&0.6303&0.6753& & 
			0.2753&0.4028&0.5000&0.5732&0.6290&0.6753 \\ 
			MIMIC&GRAM   & 
			0.5870&0.5248&0.5498&0.6024&0.6523&0.6956&
			&0.2792&0.4210&0.5211&0.5954&0.6507&0.6955\\ 
			&KAME& 
			0.5852&0.5195&0.5389&0.5873&0.6384&0.6799& & 
			0.2759&0.4164&0.5111&0.5808&0.6370&0.6799\\ 
			&MIPO& 
			\textbf{0.6446}&\textbf{0.5661}&\textbf{0.5813}&\textbf{0.6305}&\textbf{0.6752}&\textbf{0.7189}& & 
			\textbf{0.3070}&\textbf{0.4522}&\textbf{0.5502}&\textbf{0.6229}&\textbf{0.6739}&\textbf{0.7188}\\ 
			\hline
			&BRNN     & 
			0.6221&0.7011&0.7756&0.8229&0.8620&0.8845& & 
			0.5480&0.6892&0.7733&0.8226&0.8619&0.8845\\ 
			&RETAIN    & 
			0.6332&0.7124&0.7796&0.8277&0.8655&0.8907& & 
			0.5571&0.7000&0.7772&0.8274&0.8654&0.8907\\ 
			&Dipole  & 
			0.6264&0.7018&0.7696&0.8255&0.8610&0.8898& & 
			0.5514&0.6895&0.7673&0.8252&0.8609&0.8898 \\ 
			eICU&GRAM   & 
			0.6048&0.6846&0.7571&0.8101&0.8485&0.8791& &
			0.5277&0.6719&0.7549&0.8098&0.8485&0.8791\\ 
			&KAME& 
			0.6004&0.6795&0.7509&0.8093&0.8466&0.8770& & 
			0.5226&0.6668&0.7487&0.8090&0.8465&0.8770\\ 
			&MIPO& 
			\textbf{0.6848}&\textbf{0.75202}&\textbf{0.8127}&\textbf{0.8532}&\textbf{0.8847}&\textbf{0.9086}& & 
			\textbf{0.5986}&\textbf{0.73852}&\textbf{0.8102}&\textbf{0.8527}&\textbf{0.8845}&\textbf{0.9086}\\ \hline
		\end{tabular}
	}
\end{table*}

\subsection{Predictive Tasks}

The proposed model consists of two predictive tasks to simultaneously learn the integration between knowledge graph and sequential patient journey.

A task-specific predictive task is to predict the diagnosis information of the next visit. In the experiments, true labels $\bm{y}^P_t$ are prepared by grouping the ICD9 codes into 283 groups using CCS single-level diagnosis grouper\footnote{https://www.hcup-us.ahrq.gov/toolssoftware/ccs/AppendixASingleDX.txt}. This aims to improve the training speed and predictive performance, while preserving sufficient granularity for all the diagnoses. The second hierarchy of the ICD9 codes can also be used as category labels~\cite{Ma2018-gu_kame}. These two grouping methods obtain similar predictive performances.

An ontology-based disease typing task is to predict the disease category given the medical code (leaf node). The disease categories come from ``CCS\_LVL\_1'' of CCS multi-level diagnosis grouper\footnote{https://hcup-us.ahrq.gov/toolssoftware/ccs/AppendixCMultiDX.txt}, which groups the ICD9 codes into 18 categories. In the experiments, we prepared the 18 categories as true labels $\bm{y}^V_{t,i}$. This is to guarantee parent nodes learn general knowledge from their children, following the \textit{parent-child} semantic relationship. The results will be shown in Section~\ref{sec:interpret}.

\subsection{Experimental Setup}

\subsubsection{Baseline Approaches}
We compare the performance of our proposed model against the following baseline models:
\begin{itemize}
\item \textbf{Graph-based Models} include GRAM~\cite{Choi_Bahadori_2017_gram} and KAME~\cite{Ma2018-gu_kame}. They incorporate the medical ontology with an attention mechanism and recurrent neural networks for representation learning with the application to diagnosis prediction.

\item \textbf{Attention-based Models} include Dipole~\cite{Ma2017-gs_Dipole} which uses the bidirectional GRU as the backbone and assigns an attention weight for each visit and RETAIN~\cite{choi2016retain} which learns the medical concept embeddings and performs heart failure prediction via the reversed RNN with the attention mechanism.

\item \textbf{Plain RNNs} include BRNN which is the basic framework of most risk prediction models with the bidirectional GRU as the backbone.


\end{itemize}

\subsubsection{Evaluation Measures}
We measure the predictive performance by $Prec@k$ and $Acc@k$, which are defined as:

\begin{align*}
\scriptsize
    Prec@k = \frac{\texttt{\# of true positives in the top k predictions}}{\min(k,~\texttt{\# of positives})}
\end{align*}

\begin{align*}
\scriptsize
    Acc@k = \frac{\texttt{\# of true positives in the top k predictions}}{\texttt{\# of positives}}
\end{align*}

We report the average values of $Prec@k$ and $Acc@k$ and vary \textit{k} from 5 to 30 in the experiments, where $Prec@k$ aims to evaluate the coarse-grained performance, and $Acc@k$ is proposed to evaluate the fine-grained performance~\cite{Ma2018-gu_kame}. For all the measures, greater values reflect better performance.




\subsection{Results of Diagnosis Prediction}
Table~\ref{tab:Accuracy} shows both the precision and accuracy of the proposed MIPO and baselines with different \textit{k} on two real-world datasets for task-specific predictive task. From Table~\ref{tab:Accuracy}, we can observe that the performance of the proposed MIPO, in both precision and accuracy, is better than that of all the baselines on the two datasets.

On the MIMIC dataset, compared with KAME and GRAM, the precision of MIPO improves 5.94\% and 5.76\% with accuracy improving 3.11\% and 2.78\% when $k = 5$, respectively. These results suggest that it is effective to integrate medical knowledge and sequential patient journey when predicting diagnoses. Comparably, Dipole, RETAIN and BRNN do not use external knowledge in the diagnosis prediction task. Dipole and RETAIN directly learn the medical code embeddings from the input data with location-based attention mechanisms, and BRNN learns the code embeddings from the input data with bi-directional RNN. Compared with KAME and GRAM, the performances of Dipole, RETAIN and BRNN are lower, indicating that employing knowledge graph is effective with data insufficiency. However, instead of adding attention mechanisms on the past visits like Dipole and RETAIN, and simply integrating medical knowledge into visits like KAME and GRAM, the proposed MIPO aims to integrate the given knowledge graph and sequential patient journeys to improve predictive performance.

Though the number of visits and patients on the eICU is larger than that on the MIMIC dataset, the number of labels observed are much less. On this significantly insufficient dataset, MIPO still outperforms all the baselines. In the five described, Dipole, RETAIN and BRNN achieves a better performance than KAME and GRAM, which suggests that with enough data, even without external knowledge, attention-based models can still learn reasonable medical code embeddings to make accurate predictions.
However, compared with the proposed MIPO, the precision and accuracy of these three approaches are lower, which again argues that integration of medical knowledge and sequential patient journeys can improve prediction performance. 
The performance of KAME is the weakest since this approach explicitly incorporates knowledge from leaf nodes and parent nodes, which cannot adequately balance the knowledge and sequential visits. 
However, the proposed model learns the healthcare representations by taking advantage of task-specific predictive task and ontology-based disease typing task to harmoniously fuse medical knowledge and sequential patient journey.

 \begin{figure}[t]
 \centering
  \includegraphics[width=0.475\textwidth]{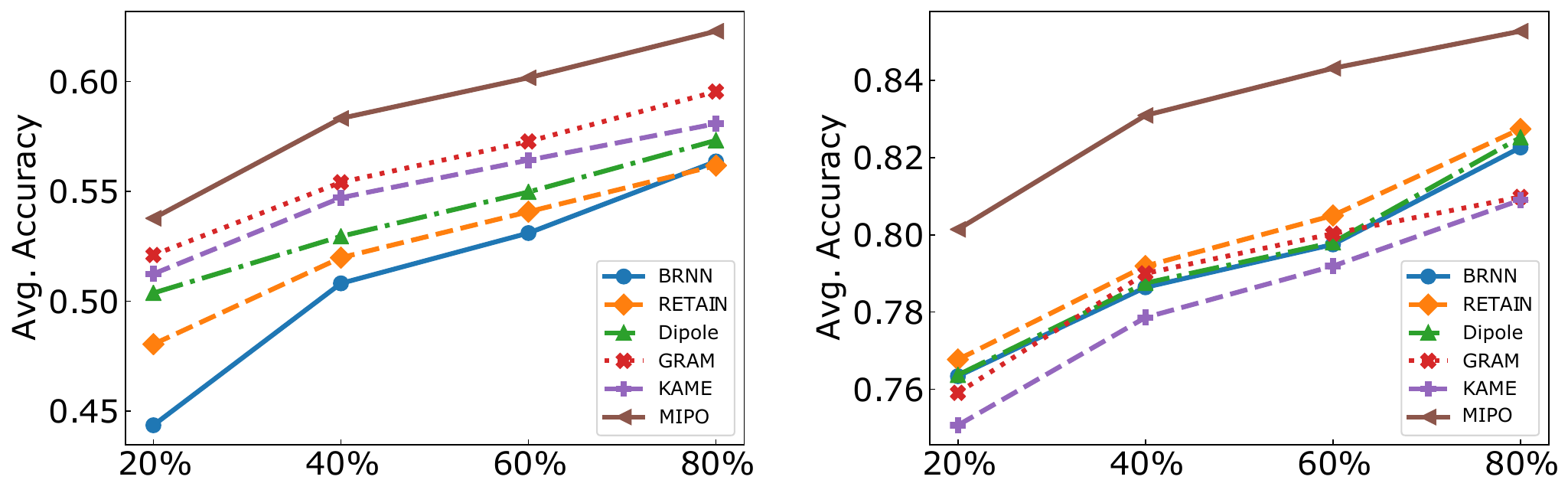}
     \begin{subfigure}[b]{0.25\textwidth}
         \centering
         \caption{MIMIC}\label{sub-fig:mimic}
     \end{subfigure}
     \hfill
     \begin{subfigure}[b]{0.22\textwidth}
         \centering
         \caption{eICU}\label{sub-fig:eicu}
     \end{subfigure}
  \caption{Acc@20 of diagnoses prediction on MIMIC and eICU, size of training data is varied from 20\% to 80\%.}
  \label{fig:mimic-eicu}
\end{figure}

 \begin{figure*}
 \centering
  \includegraphics[width=0.8\textwidth]{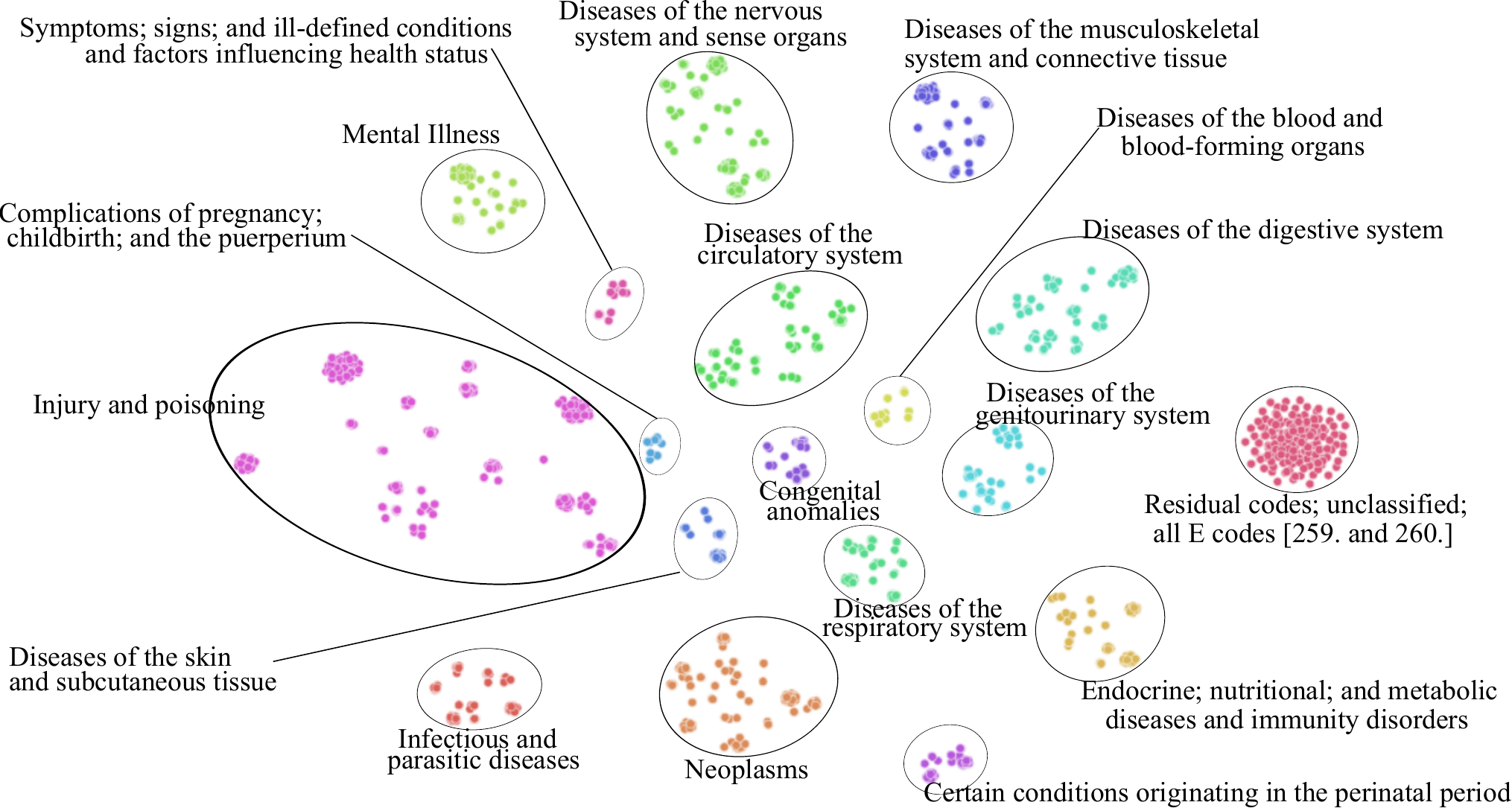}
  \caption{Annotations of MIPO Diagnosis Embedding }
  \label{fig:annot}
\end{figure*}

 \begin{figure*}
 \centering
  \includegraphics[width=0.8\textwidth]{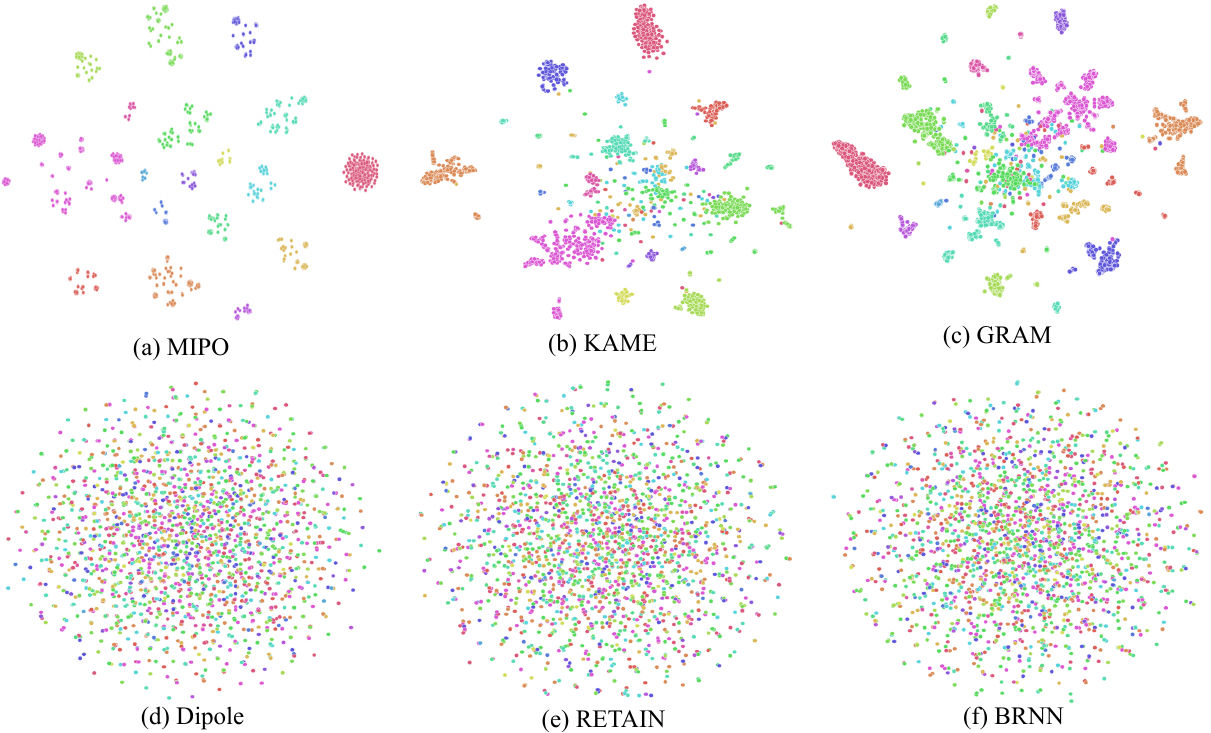}
  \caption{\textit{t}-SNE Scatterplots of Medical Codes Learned by Predictive Models on the MIMIC dataset.}
  \label{fig:t-sne}
\end{figure*}

\subsection{Data Sufficiency Analysis}

In order to analyze the influence of data sufficiency on the predictions, we conduct the following experiments on the MIMIC and eICU datasets, respectively.  We randomly split the data into training set, validation set and test set, and fix the size of the validation set at 10\%. To validate robustness against insufficient data, we vary the size of the training set to form four groups: 20\%, 40\%, 60\% and 80\%, and use the remaining part as the test set. The training set in the 20\% group is the most insufficient for training the proposed and baseline models, while the data in the 80\% group are the most sufficient for training models. Finally, we calculate the accuracy of labels in each group. Figures~\ref{fig:mimic-eicu} show the Acc@20 on both the MIMIC and eICU datasets. Note that similar results can be obtained when $k$ = 5, 10, 15, 25 or 30.

From Figure~\ref{fig:mimic-eicu}, we can observe that the accuracy of the proposed MIPO is higher than that of baselines in all groups on both MIMIC and eICU datasets. 
KAME and GRAM achieve better performances on MIMIC than other approaches on MIMIC, which shows that, with insufficient data, KAME and GRAM still learn reasonable medical code embeddings and improve predictions. The performance of BRNN in the groups 20\%, 40\%, 60\% is the worst since this approach does not use any attention mechanism or external knowledge. 
When the training data on the eICU dataset is significantly insufficient, the proposed MIPO still significantly outperforms baselines in all groups.
We observe that the performance obtained by the models using medical knowledge remains approximately the same (GRAM) or even drop (KAME). 
The underlying reason may be that KAME and GRAM over-fit the insufficient data using the medical knowledge. 
Thus, the models learn larger weighting for knowledge than with sequential visits. 
Furthermore, as shown in Figure~\ref{sub-fig:eicu}, the average accuracy of Dipole, RETAIN and BRNN is better than that of both KAME and GRAM, indicating that information of sequential visits plays a more important role under insufficient data. 
These observations can also be found in Table~\ref{tab:Accuracy}. 
It is again to demonstrate that the proposed MIPO harmoniously balances medical knowledge and patient journeys when the EHRs data is insufficient.


\subsection{Interpretable Representation Analysis}\label{sec:interpret}

To qualitatively demonstrate the interpretability of the learned medical code representations by all the predictive models on the MIMIC dataset, we randomly select 2000 medical codes and then
plot on a 2-D space with t-SNE~\cite{maaten2008visualizing} 
shown in Figures~\ref{fig:annot} and \ref{fig:t-sne}. Each dot represents a diagnosis code, and their color represents the disease categories while the text annotations represent the detailed disease categories in CCS multi-level hierarchy. 

From Figure~\ref{fig:annot}, we can observe that MIPO learns interpretable disease representations that are in accord with the hierarchies of the given knowledge graph $\mathcal{G}$, and obtains 18 non-overlapping clusters. 
As shown in Figure~\ref{fig:t-sne}, KAME and GRAM learn reasonably interpretable disease representations for partial categories, as there is large number of dots over-lapping in the centers of Figures~\ref{fig:t-sne}b and \ref{fig:t-sne}c.  Figures~\ref{fig:t-sne}d, \ref{fig:t-sne}e and \ref{fig:t-sne}f confirm that without a knowledge graph, simply using the co-occurrence or supervised predictions cannot easily provide for learning interpretable representations. In addition, the predictive performance of MIPO is much better than that of KAME and GRAM, as shown in Table~\ref{tab:Accuracy}, which proves that the proposed model does not affect the interpretability of medical codes. Moreover, it significantly improves the prediction accuracy.

\section{Conclusions}\label{sec:Con}
This paper presents MIPO, a novel approach that integrates medical knowledge and the patient journey to enhance healthcare representation. By introducing a knowledgeable encoder and employing two predictive tasks—sequential diagnosis prediction and disease category classification—MIPO effectively captures heterogeneous information from both patient records and a knowledge graph. Empirical evaluations on two real-world medical datasets reveal that MIPO not only outperforms baseline methods but also remains robust in scenarios with limited data. Moreover, the visualizations of medical code representations offer insights into MIPO’s interpretability, underscoring its potential for real-world healthcare applications.

\bibliographystyle{IEEEtran}
\bibliography{main}

\end{document}